\documentclass[10pt, a4paper]{article}

\usepackage{lrec-coling2024} 

\usepackage{xurl}
\usepackage{times}
\usepackage{array}
\usepackage{amsmath}
\usepackage{ragged2e}
\usepackage{multicol}
\usepackage{multirow}
\usepackage{latexsym}
\usepackage{graphicx}
\usepackage{adjustbox}
\usepackage{microtype}
\usepackage[T1]{fontenc}
\usepackage[utf8]{inputenc}

\usepackage{titlesec}
\titlespacing{\section}{0pt}{2ex}{1ex}
\titlespacing{\subsection}{0pt}{1.5ex}{0.5ex}
\titlespacing{\subsubsection}{0pt}{1ex}{0.5ex}

\title{Uncovering Agendas: A Novel French \& English Dataset \\for Agenda Detection on Social Media}

\name{Gregorios A Katsios*, Ning Sa**, Ankita Bhaumik*, Tomek Strzalkowski**} 

\address{*Department of Computer Science, **Department of Cognitive Science, \\
         Rensselaer Polytechnic Institute, \\
         \{katsig, san2, bhauma, tomek\}@rpi.edu\\}

\abstract{
The behavior and decision making of groups or communities can be dramatically influenced by individuals pushing particular agendas, e.g., to promote or disparage a person or an activity, to call for action, etc.. In the examination of online influence campaigns, particularly those related to important political and social events, scholars often concentrate on identifying the sources responsible for setting and controlling the agenda (e.g., public media). In this article we present a methodology for detecting specific instances of agenda control through social media where annotated data is limited or non-existent. By using a modest corpus of Twitter messages centered on the 2022 French Presidential Elections, we carry out a comprehensive evaluation of various approaches and techniques that can be applied to this problem. Our findings demonstrate that by treating the task as a textual entailment problem, it is possible to overcome the requirement for a large annotated training dataset.
 \\ \newline }

\begin{document}

\maketitleabstract

\section{Introduction}
An agenda, being a collection of items to be attended to in a certain order, can have a significant impact on the actions of a group, especially in the context of interpersonal communication and relationships. In human communication, an agenda refers to the underlying intentions or motives of a particular person or group to steer the conversation in a particular direction in order to achieve a desired effect \citep{PoliticalAgendaSeptember2023,HiddenAgendaSeptember2023,AgendaJuly2023}. The individuals who establish and direct the agenda often exercise considerable control and influence over their audience. In the sociolinguistics of group behavior, the concept of agenda control is widely recognized as a strong indicator of both leadership and influence, as evidenced by numerous studies in the field \citep{wang2018glue,broadwell2013modeling,strzalkowski2013influence}.

When studying the impact of online influence campaigns, such as those surrounding significant political and social events (e.g., elections), researchers often focus on evidence of agenda-setting activity emanating from particular sources. These sources could be traditional public media or clandestine online groups that wish to shape public opinion. According to social science literature, there are three distinct levels of agenda setting. At level one, the public is told explicitly what to think and do in a given situation, for example, to vote for a particular candidate. In the second level of agenda setting, rather than prescribing specific beliefs or actions, the influencers emphasize certain aspects of their targets (e.g., political candidates) as either positive or negative, outwardly leaving the public to form their own opinions \citep{mccombs1997candidate,balmas2010candidate,meraz2011fight}. At the third level, multiple targets are associated to one another through direct comparison or juxtaposition \cite{guo2012expanded} thus imparting apparent preferences onto the public.

In this study, we are interested in both level one and level two agenda setting activities and how to detect their presence in social media, with a specific focus on the 2022 French Presidential Elections. Our goal is to detect specific instances of agendas being actively promoted via social media messaging (see Table 1). Our focus is on Twitter messages (tweets), including retweets, replies, and quotes, posted in multiple languages during the relevant time period. The objective is to automatically tag each tweet with appropriate agenda labels, with each label representing a type of agenda, not necessarily explicit, that may arise in a political context.

The agenda labels under consideration and their definitions in English are presented in Table \ref{tab:definitions}. In this work, we focus on a set of "call for action" agendas which seek to inspire concrete action(s). The labels were curated by political science experts, and the curation process is outside of the scope of this paper; however, the agenda labels, which we discuss in more details below, were chosen to apply on most election-style events, although not necessarily to other events such as international conflicts. Given the novelty and an ad-hoc nature of the agenda labeling problem, the lack of pre-existing annotated training data, and the practical limitations of obtaining sufficient quantities of such data (which is true of most real-world applications), our approach focuses on utilizing small, expert annotated samples and relying on zero-shot and few-shot methods. Our proposed framework provides a general solution for classifying social media messages that operates on a set of ad-hoc agenda labels.

\begin{table*}[!hbt]
    \centering
    \begin{tabular}{p{0.2\linewidth} | p{0.74\linewidth} }
        \hline
        \textbf{Agenda Labels} & \textbf{Agenda Definitions (EN)} \\
        \hline
        Online Solidarity & The message encourages readers to share information relevant to a cause, promote or magnify the positions of specific individuals, use symbols or language in online profiles to demonstrate support for a specific position on an issue. \\
        \hline
        Engagement & The message encourages readers to engage in the formal political process, either by voting, attending public government meetings, assemblies, etc., to support or oppose a candidate, party, law, political position, or (nominally) collective action by a government. \\
        \hline
        Disengagement & The message encourages readers to disengage from a normal political, economic, or social process in order to demonstrate opposition to the status quo on a specific issue or to highlight the importance of a specific stance. \\
        \hline
        Peaceful Protest & The message encourages readers to protest peacefully, to attend rallies, marches, and other forms of mass political demonstration, etc. in support of or opposition to a cause. The action or demonstration urged by the document must be non-violent in nature. \\
        \hline
        Violent Action & The message encourages readers to engage personally in violent or destructive action (bombing, destruction of property, formation of militias, fighting in foreign countries in a mercenary capacity, etc.).\\
        \hline
        Other & The text is about something else. \\
        \hline
    \end{tabular}
    \caption{Agenda labels with definitions.}
    \label{tab:definitions}
\end{table*}

Since an agenda is defined as an intention, explicit or not, behind a message (or a set of messages), it is reasonable to assume that the message \textit{implies} the agenda, or at least it is meant to imply (assuming the message is understood). In other words, if we can show that a message implies one of the agendas in our agenda collection, we can assign the corresponding agenda label to this message.

Accordingly, we propose to cast the agenda detection task as a textual entailment problem. In previous studies \citep{yin2019benchmarking}, text classification has been viewed through the lens of textual entailment with promising results. This approach imitates how humans make decisions while annotating text examples, picking the correct label among all possible labels. Human annotators are often given a task description, as well as label definitions that explain the meaning of each candidate label. Equipped with these definitions, a human can understand the problem and mentally construct a hypothesis by picking a candidate label to fill in the blank: \textit{"This text is about \_\_\_"}. Then they ask themselves if this hypothesis is true given the text example.

We treat agenda detection as a textual entailment problem so that our model can gain knowledge from entailment datasets \citep{bowman2015large,williams2017broad,dagan2006pascal,bentivogli2009fifth}. We should mention that the textual entailment approach is well suited for agenda detection, and our approach is not limited to a predefined set of labels as it can be extended to arbitrary sets using generative methods.

\section{Related Work}
Recent studies examine methods to detect influence indicators from individual social media messages. For instance, \citet{bhaumik2023adapting} identify emotions expressed by the authors of social media posts in relation to political issues or causes. To explore the topic of agenda detection in social media, we present a review of selected literature in the domains of traditional agenda detection and text classification through the lens of textual entailment, as our proposed models draw inspiration and incorporate elements from these fields. 

\subsection{Agenda Detection}
The impact of various agendas being pushed through the media (both official and unofficial) have on shaping public opinion has been widely studied, as has the interplay between the news outlets and the social media. For example, \citet{mccombs1997candidate} attempts to understand how media agendas shape or influence the public’s opinion on political candidates, and \citet{vargo2014network} studies how the public selectively accepts media agendas. Additionally, the effect that news media and social media have on each other is closely examined in \citep{su2020delineating}.

In the absence of large annotated datasets, scholars often perform manual analysis to detect agendas. \citet{mccombs1997candidate} hand-coded news articles and surveys with attribute labels signaling the candidates’ ideology and positions on public issues, their qualifications, experience, their personal characteristics and personality. Similarly, \citet{su2019agenda} conducted a manual analysis of a sample of collected tweets, labeling them with 11 distinct classes on the topic of Climate Change. Automated methods such as those used in \citep{vargo2014network}, \citep{ceron2016first} and \citep{haim2018sets} utilized sets of keywords to detect topics and sentiment associated with the target agendas, rather than the agendas themselves. 

More recently, several studies explored machine learning methods for the detection of agendas in big data. In \citep{su2020delineating,su2022networked,guo2019media}, the authors first utilize topic modeling to identify the topics within their datasets. Then, human experts manually develop agenda labels associated with each topic. Subsequently, multiple annotators tag a subset of the data using the agenda labels developed in the previous step. The labeled data is then used to train a set of Support Vector Machine (SVM) \citep{boser1992training} classifiers, one per each agenda label. If the average performance of the classifiers was not satisfactory (e.g., F1 < 0.7), more data was annotated and the training was repeated. 

\citet{guo2019media} collected and annotated 2000 news articles with 31 topic labels for the task of detecting inter-media agenda setting in Chinese online news. In \citep{su2022networked}, the authors explored the information flow between newspaper and Twitter focusing on the topic of Black Lives Matter. A total of 1500 news articles and 5000 tweets were annotated with 16 topic labels and 5 affect labels. \citet{su2020delineating} analyzed the Hong Kong Movement and annotated 3000 tweets and 500 news articles with 3 affect labels: \textit{pro-protest, neural, anti-protest}, and 13 topic labels: \textit{Violence of police}, \textit{UK politics},\textit{ US politics}, \textit{Sino-US relation}, \textit{HK legislation}, \textit{Violence of protesters}, \textit{HK economy}, \textit{Overseas Chinese students}, \textit{Democracy \& human rights}, \textit{HK-Mainland relations}, \textit{Public security} and \textit{Social media \& Entertainment}.

The authors of \cite{chen2019top} used a similar approach, but deployed different classifiers, trained on 2500 annotated microblog messages centered on the topic of Chinese Nationalism on Social Media. We note that all the above approaches are costly and impractical, particularly in novel and rapidly evolving situations.

\subsection{Textual Entailment Text Classification}
In their work, \citet{yin2019benchmarking} introduced a framework for text classification by formulating it as a series of premise-hypothesis pairs, where the premise is the text to be classified and the hypotheses represents the candidate labels, essentially transforming the problem into a textual entailment challenge. They demonstrated the efficacy of this method, and released a benchmark dataset for zero-shot text classification. Subsequently, this approach has been widely adopted and expanded for many zero-shot text classification tasks \citep{shu2022zero,zhang2020discriminative,wang2021entailment,seoh2021open}. Our work builds upon this basic methodology by applying textual entailment to the task of agenda detection, and we evaluate and compare various approaches that could also be used to solve this problem, including conventional text classification methods. By doing so, we demonstrate the utility of this framework for addressing the task of agenda detection in the absence of large annotated datasets.

\section{Data}
Our proposed model makes use of pre-existing textual entailment datasets, described in the following paragraphs, to gain general knowledge. This pre-training methodology exposes the model to a multitude of linguistic and factual scenarios. Through the utilization of textual entailment datasets, the model learns to make sense of contradictions, inferences, and entailment relationships. Then, the model is trained on agenda specific data. 

\subsection{Pre-training Data}
To teach our model how to solve the textual entailment task, we deploy three widely used datasets into an early fine-tuning training step. These datasets are i) the Stanford Natural Language Inference (SNLI) dataset \citep{bowman2015large}, ii) the Multi-genre Natural Language Inference (MNLI) dataset \citep{williams2017broad}, and iii) the Recognizing Textual Entailment (RTE) dataset \footnote{RTE dataset: \url{https://dl.fbaipublicfiles.com/glue/data/RTE.zip}} \cite{dagan2006pascal,bentivogli2009fifth}, which is also part of the GLUE benchmark \citep{wang2018glue}. 

We convert all datasets to represent binary classification problems, where for three-class datasets, we collapse neutral and contradiction into not entailment, so that our model learns to distinguish entailment from not entailment. All three datasets come with predefined training examples, so we merge all three training partitions into a single training set. Since our downstream task of agenda detection contains messages that are in English and French, we automatically translate\footnote{Machine Translation model: \url{https://huggingface.co/Helsinki-NLP/opus-mt-en-fr}} 30\% of the combined SNLI/MNLI/RTE training dataset from English to French.

In our experiments, discussed further in this article, we test whether including this data collection into the fine-tuning process improves the performance of the textual entailment approach.

\subsection{Fine-tuning Data} 
To fine-tune our model to the task of agenda detection, we use the publicly available dataset of tweets\footnote{Un-tagged Twitter corpus: \url{https://www.kaggle.com/datasets/jeanmidev/french-presidential-online-listener}}. This collection of Twitter messages contains posts on the topic of the 2022 French Presidential Elections that were made on the platform between 12 November, 2021 and 30 April, 2022. The posts, primarily written in French, were filtered using keywords including the candidate names and their associated official Twitter account, however, they do not have any agenda annotations. Since these messages were collected prior to 2023, the length of each post is limited to 280 characters.


We bootstrap an agenda training dataset by leveraging a multi-lingual sentence embedding model\footnote{Multi-lingual sentence embedding model: \url{https://huggingface.co/sentence-transformers/paraphrase-multilingual-mpnet-base-v2}} \citep{reimers-2020-multilingual-sentence-bert}. We implement an automatic labeling procedure for tweets by ranking messages using cosine similarity scores computed between the agenda definition embeddings (shown in Table \ref{tab:definitions}) and the embedding of each tweet. To ensure representation of all agenda classes in the final dataset, we gather a minimum of 500 messages with the highest similarity score for each class, automatically assigning the corresponding label. It should be noted that the number of messages retrieved for each agenda class varies, with some classes yielding more messages than others. Table \ref{tab:tw_examples} contains example messages for each of the agenda labels in French and English.



\begin{table*}[!hbt]
    \centering
    \begin{tabular}{p{0.2\linewidth} | p{0.74\linewidth} }
        \hline
        \textbf{Agenda Labels} & \textbf{Example Messages} \\
        \hline
        \multirow{2}{*}{Online Solidarity}
        & \textbf{Fr:} @JLMelenchon Pour dégager tout ça, pour \#MacronDestitution et \#RefonderLaSociété, soutenez directement cette action concrète (likez et partagez directement le tweet principal). Et surtout, signez la pétition! Merci!\\
        \cline{2-2}
        & \textbf{En:} @JLMelenchon To clear all this, for \#MacronDestitution and \#RefoundThe Company, support this concrete action directly (like and share the main tweet directly). And above all, sign the petition! Thank you! \\
        
        \hline
        \multirow{2}{*}{Engagement}
        & \textbf{Fr:} Chaque vote compte. Allez voter au nom d’un citoyen. \\
        \cline{2-2}
        & \textbf{En:} Every vote counts. Go vote on behalf of a citizen. \\
        
        \hline
        \multirow{2}{*}{Disengagement}
        & \textbf{Fr:} @montebourg @JLMelenchon Ça a changé les élections présidentielles, une bonne raison pour pas aller voter \\
        \cline{2-2}
        & \textbf{En:} @montebourg @JLMelenchon It changed the presidential elections, a good reason not to go vote \\
        
        \hline
        \multirow{2}{*}{Peaceful Protest}
        & \textbf{Fr:} @AurelieM0813 @f\_philippot Faire monter le niveau de conscience par tout les moyens de reinformation possibles, manifester en masse pacifiquement, boycotter.. tenir bon jusqu'à ce que leur château de cartes s'effondre.. sinon je ne sais pas \\
        \cline{2-2}
        & \textbf{En:} @AurelieM0813 @f\_philippot Raise the level of awareness by all possible means of reinformation, demonstrate peacefully, boycott.. hold on until their house of cards collapses.. otherwise I don't know  \\
        
        \hline
        \multirow{2}{*}{Violent Action}
        & \textbf{Fr:} Je vais utiliser la violence pour combattre la violence. Comme ça parce que c'est moi le plus violent les autres ils arrêteront d'être violent \#LogiqueInfaillible \#TraduisonsLes \\
        \cline{2-2}
        & \textbf{En:} I will use violence to fight violence. Like that because I'm the most violent, the others will stop being violent \#LogiqueInfaillible \#TraduisonsLes \\
        
        \hline
        \multirow{2}{*}{Other}
        & \textbf{Fr:} Ce prof retraité en phase avec le fantasme macroniste. \\
        \cline{2-2}
        & \textbf{En:} This retired teacher in tune with the macronist fantasy \\
        \hline
    \end{tabular}
    \caption{Agenda labels with example messages.}
    \label{tab:tw_examples}
\end{table*}


Following the initial retrieval process, two human annotators independently examine the messages to confirm their alignment with the automatically assigned label, and when necessary, they reassign the appropriate agenda class labels. In instances where a message cannot be categorized into any of the agenda classes, it is designated as "Other". Any discrepancies or disagreements that arise during the annotation process are addressed through discussion and consensus between the annotators. After the two annotators annotate the tweets, we calculate the Inter Rater Reliability. The annotators achieve 97.5\% of agreement and Cohen's Kappa of 0.89. 


\begin{table}[hbt!]
    \centering
    \begin{tabular}{c | c <{\hspace{-3pt}} | c <{\hspace{-3pt}} | c <{\hspace{-3pt}} | c <{\hspace{-3pt}}}
        \hline
        \textbf{Agenda Labels} & \textbf{Total} & \textbf{Train} & \textbf{Dev} & \textbf{Test}  \\
        \hline
        Online Solidarity & 97 & \multirow{7}{*}{80\%} & \multirow{7}{*}{10\%} & \multirow{7}{*}{10\%} \\ \cline{1-2}
        Engagement        & 120   &       &     &      \\ \cline{1-2}
        Disengagement     & 120   &       &     &      \\ \cline{1-2}
        Peaceful Protest  & 108   &       &     &      \\ \cline{1-2}
        Violent Action    & 96    &       &     &      \\ \cline{1-2}
        Other             & 506   &       &     &      \\ \cline{1-2}
        Total             & 1012  &       &     &      \\ \hline
    \end{tabular}
    \caption{Number of posts per class in the agenda dataset.}
    \label{tab:split}
\end{table}

The human annotation process yields a varying number of messages per class, with quantities ranging from 96 to 120, as detailed in Total column of Table \ref{tab:split}, except for the "Other" class for which we randomly select 506 messages (equivalent to the cumulative sum of the other five classes). Consequently, our final multi-label dataset contains a total of 1012 annotated messages, 35 of which have more than one labels. 

The dataset includes 10 original English tweets and 1002 original French messages. Utilizing the Google Translation API, each English message was translated into French, and vice versa. Therefore, our final dataset contains 1012 English and 1012 French texts. For more details regarding the agenda dataset and the annotation process, please refer to the Appendix \ref{sec:appendix_b}.

To facilitate the training and evaluation of our models, we establish three training, development (dev) and testing sets, such that the dev and test sets are non-overlapping. We create the dev and test sets by taking a 10\% random sample of the total number of messages. This data collection is carefully designed to be used in both textual entailment and traditional text classification methodologies, which we employ as our baseline. We make our annotated dataset and code available for public use so that it can be of benefit to future research \footnote{GitHub Repository: \url{https://github.com/HiyaToki/Uncovering-Agendas/}}.


\section{Method}
As explained earlier, we cast the problem of agenda classification as a textual entailment problem. This enables our system to gain further knowledge from entailment datasets, essentially learning how to imitate the human decision-making process of categorizing text.

Following similar methods to \citep{yin2019benchmarking}, we depart from the traditional text classification methods where labels are denoted as indices and models lack any understanding of their specific interpretation or meaning. Instead, the labels are transformed into a set of natural language hypotheses that the input messages will be paired against and the truth value of the label can be decided. This way the system can understand the described task and the meaning of the labels by associating the input text and the context of the hypotheses.

\subsection{Models}
Our proposed approach leverages the T5 \citep{raffel2020exploring} language model and its variants, mT5\footnote{mT5: \url{https://huggingface.co/google/mt5-base}} \citep{xue2020mt5} and T5v1.1\footnote{T5 v1.1: \url{https://huggingface.co/google/t5-v1_1-base}}. T5 stands out for its exceptional performance, owing to a number of key factors, such as its encoder-decoder architecture, the corrupting span denoising objective, and the utilization of an extensive pre-training dataset. Furthermore, T5v1.1 and mT5 are further enhanced by the integration of GeLU \citep{shazeer2020glu} activation. mT5, in particular, has been pre-trained on over 120 languages, including French, which is of particular interest in the context of our task.

For our baselines models, we use BERT\footnote{BERT: \url{https://huggingface.co/bert-base-uncased}}, mBERT\footnote{mBERT: \url{https://huggingface.co/bert-base-multilingual-uncased}}, an English pretrained Sentnece Transformer \footnote{SBERT: \url{https://huggingface.co/sentence-transformers/all-mpnet-base-v2}}, as well as a muilti-lingual pre-trained Sentence Transformer\footnote{mSBERT: \url{https://huggingface.co/sentence-transformers/paraphrase-multilingual-mpnet-base-v2}}.

For our training processes, we use the following hyper-parameter settings for all Transformer-based models: Batch Size = 32, Epochs = 5, Weight Decay = 0.01, and Warm-up Ratio = 0.1. For the T5-based models we use a Learning Rate of 1e-4, while for BERT-based models we use a Learning Rate of 2e-5.

\subsection{Pre-training}
For models trained under the Textual Entailment framework, we first train using the binarized SNLI/MNLI/RTE pre-training dataset discussed above, such that our models learn to distinguish entailment versus not entailment when a premise and a hypothesis are given as inputs. This step has been shown \citep{yin2019benchmarking} to improve the robustness of the model on zero-shot text classification tasks. For traditional classification approaches, we do not include a similar pre-training step.

\subsection{Fine-tuning}
To adapt our agenda dataset into a format suitable for Textual Entailment, we convert each unique agenda label into a hypothesis following the process described in the following section. Then, we consider each input message as a premise that has positive hypotheses (entailment) corresponding to the ground truth label, while negative labels provide negative hypotheses (not entailment). During fine-tuning, we form all possible positive (i.e., entailment) message-to-label examples (where a message may be associated with more than one agenda), and in addition, two negative examples (i.e., non-entailment) for each positive. This mimics the distribution of entailment/not entailment found in our pre-training dataset. For traditional classification approaches, each message is associated with its ground truth labels in a multi-label fashion.

\begin{table*}[hbt!]
    \centering
    \begin{tabular}{p{0.2\linewidth} | p{0.35\linewidth} | p{0.35\linewidth}}
        \hline
        \textbf{Agenda Labels} & \textbf{Hypotheses (EN)} & \textbf{Hypotheses (FR)} \\
        \hline
        Online Solidarity & The author encourages readers to share information relevant to a cause, promote the positions of individuals and show support for a position on an issue. & L'auteur encourage les lecteurs à partager des informations pertinentes pour une cause, à promouvoir les positions des individus et à montrer leur soutien à une position sur une question. \\
        \hline
        Engagement & The document encourages readers to engage in the formal political process, by voting and attending public government meetings , to support or oppose a candidate, party, law or a political position. & Le document encourage les lecteurs à s'engager dans le processus politique formel, en votant et en assistant aux réunions publiques du gouvernement, pour soutenir ou s'opposer à un candidat, un parti, une loi ou une position politique. \\
        \hline
        Disengagement & The author wants the readers to disengage from a normal political process in order to demonstrate opposition to the status quo on an issue or to highlight the importance of a stance. & L'auteur souhaite que les lecteurs se désengagent d'un processus politique normal afin de manifester leur opposition au statu quo sur une question ou de souligner l'importance d'une position. \\
        \hline
        Peaceful Protest & The message motivates the readers to protest peacefully in support of or opposition to a cause. & Le message motive les lecteurs à manifester pacifiquement pour soutenir ou s'opposer à une cause. \\ 
        \hline
        Violent Action & The author rallies the audience to engage personally in violent or destructive action. & L'auteur rallie le public à s'engager personnellement dans une action violente ou destructrice. \\ 
        \hline
        Other & The text is about something else. & Le texte parle d'autre chose. \\
        \hline
    \end{tabular}
    \caption{Agenda labels with hypotheses.}
    \label{tab:hypotheses}
\end{table*}

\subsubsection{Generating Hypotheses from Agenda Labels}
An integral part of our approach is the construction of hypotheses representing the agenda classes. Here, we use the definition of each class, see Table \ref{tab:definitions}, as a guide to write succinct hypotheses in natural language. We first write the hypotheses in English and use machine translation to obtain their French versions. The hypotheses we used in our experiments are listed for each class in Table \ref{tab:hypotheses}.


\subsubsection{Interpreting Agenda Predictions from Textual Entailment}
Finally, textual entailment classification results can be interpreted into one or more agenda classes. As our base case, when the model predicts non-entailment for all possible hypotheses for an input example, we resolve it as the "Other" agenda class and output it as the final prediction for that example. In the cases where there are one or more entailment predictions for some input text, all agenda classes corresponding to those hypotheses form the final output prediction.

When yielding confidence scores, for each class we look at the probability of the corresponding hypothesis being entailed. For generative models, we use the probability of the related token. 

\section{Experimental Set-up \& Results}
Our textual entailment based approach is evaluated against a range of baseline techniques, such as conventional text classification and semantic search. We adopt a multi-class multi-label approach evaluation process, as our textual entailment approach predicts the entailment of a premise (tweet message) with respect to each of the hypotheses, one class at a time.

Each model is trained and evaluated three times, once for each individual training and corresponding testing sets. After training the models, we perform the evaluations by varying the decision threshold for each model. This threshold adjustment is based on the weighted average F1-score. This evaluation method is cost-effective as it can be performed after obtaining predictions on the test set without updating the underlying model. The decision threshold value reflects the model's confidence level, with higher values indicating greater confidence and lower values indicating that predictions with low confidence are accepted, which can result in increased False Positives. We set the minimum possible threshold to 0.3, as we are not interested in trivial scenarios where predictions include all of the available agenda labels. In the sections below, we report the averaged F1-scores across the three runs, and their standard deviation. In the Appendix \ref{sec:appendix_c}, we present detailed results for each of the three runs. 

\begin{table*}[hbt!]
    \centering
    \begin{tabular}{p{0.13\linewidth} | p{0.22\linewidth} | p{0.05\linewidth} | p{0.05\linewidth} | p{0.07\linewidth} || p{0.05\linewidth} | p{0.05\linewidth} | p{0.07\linewidth}}
        \hline
        \multicolumn{2}{c |}{\multirow{2}{*}{\textbf{Models}}} & \multicolumn{3}{c ||}{\textbf{AVG}} & \multicolumn{3}{ c }{\textbf{STDEV}} \\ 
        \cline{3-8}
        \multicolumn{2}{c |}{} & \textbf{EN} &\textbf{FR} & \textbf{Overall} & \textbf{EN} & \textbf{FR} & \textbf{Overall} \\
        \hline
        \multirow{4}{=}{\textbf{Semantic Search}} & hypotheses-SBERT  & 0.21 & 0.43 & 0.37 & 0.04 & 0.02 & 0.01 \\
                                                  \cline{2-8}
                                                  & hypotheses-mSBERT & 0.24 & 0.38 & 0.32 & 0.03 & 0.02 & 0.01 \\
                                                  \cline{2-8}
                                                  & labels-SBERT      & 0.17 & 0.08 & 0.14 & 0.01 & 0.06 & 0.02  \\
                                                  \cline{2-8}
                                                  & labels-mSBERT     & 0.18 & 0.18 & 0.18 & 0.03 & 0.04 & 0.03 \\
        \hline
        \multirow{5}{=}{\textbf{Pre-trained}} & rte-en-BERT  & 0.38 & 0.43 & 0.42 & 0.03 & 0.03 & 0.03 \\
                                              \cline{2-8}
                                              & rte-bi-mBERT & 0.37 & 0.39 & 0.38 & 0.05 & 0.04 & 0.04 \\
                                              \cline{2-8}
                                              & rte-en-T5    & \textbf{0.48} & \textbf{0.46} & \textbf{0.48}$^*$ & 0.01 & 0.01 & 0.01 \\
                                              \cline{2-8}
                                              & rte-bi-mT5   & 0.38 & 0.44 & 0.42 & 0.02 & 0.03 & 0.02 \\
                                              \cline{2-8}
                                              & mnli-BART    & 0.33 & 0.43 & 0.40 & 0.01 & 0.05 & 0.03 \\
        \hline
    \end{tabular}
    \caption{zero-shot evaluation results. We report the averaged F1-scores across three runs. An "m" in the models name indicates the use of the multi-lingual underlying model, while "rte-en" and "rte-bi" refer to the use of English-only or bi-lingual pre-training using our combined RTE dataset. The * indicates overall results that are statistically significant better than all others at $\alpha$ = 0.05}
    \label{tab:0_shot_results}
\end{table*}

\subsection{Zero-shot Agenda Detection}
In this experiment, we evaluate the performance of our proposed model for detecting agendas in social media on the zero-shot setting. To provide a comprehensive evaluation, we compare our model with several baselines, including BERT \citep{devlin2018bert}, SBERT \citep{reimers-2019-sentence-bert} and mnli-BART\footnote{mnli-BART: \url{https://huggingface.co/facebook/bart-large-mnli}}. 

Our semantic search baselines use SBERT to obtain sentence embeddings of the messages and the hypotheses, but we also test a variant using the label's text. Then, we compare the message embeddings to each hypothesis embedding using cosine-similarity. The computed score serves as the confidence that the message belongs to the agenda class specified by each hypothesis. This baseline approach yields four models, two comparing the English-only (all-mpnet-base-v2) versus the multilingual model (paraphrase-multilingual-mpnet-base-v2), and two comparing the use of hypotheses versus the labels themselves.

For our proposed approach in the zero-shot setting, we pre-train the models on the combination of the SNLI, MNLI, and RTE datasets. We compare the performance of the T5 model with BERT and pre-train on either the English-only or bilingual version of the combined RTE dataset. For instance, the "t5-v1.1-base" model is fine-tuned using the English-only version, while the "mt5-base" model is fine-tuned using the bilingual version. The models are then applied directly to the agenda test sets to generate predictions.

The availability of in-domain data is critical for training a robust classification model. Our zero-shot evaluation results (Table \ref{tab:0_shot_results}) show that a lack of in-domain data leads to lower model performance. However, the models pre-trained on our combined RTE dataset using the textual entailment framework exhibit significant improvements over the Semantic Search baselines, with an overall F1-score of 0.48, demonstrating the potential of this approach. Our best performing 0-shot model, rte-en-T5, is statistically significantly better than the all the Semantic Search models at significance level $\alpha$ = 0.01. When we relax the significance level at $\alpha$ = 0.05, then \textit{rte-en-T5} is also significantly better than all other 0-shot pre-trained models, including mnli-BART.

\begin{table*}[hbt!]
    \centering
    \begin{tabular}{p{0.13\linewidth} | p{0.22\linewidth} | p{0.05\linewidth} | p{0.05\linewidth} | p{0.07\linewidth} || p{0.05\linewidth} | p{0.05\linewidth} | p{0.07\linewidth}}
        \hline
        \multicolumn{2}{c |}{\multirow{2}{*}{\textbf{Models}}} & \multicolumn{3}{c ||}{\textbf{AVG}} & \multicolumn{3}{ c }{\textbf{STDEV}} \\ 
        \cline{3-8}
        \multicolumn{2}{c |}{} & \textbf{EN} &\textbf{FR} & \textbf{Overall} & \textbf{EN} & \textbf{FR} & \textbf{Overall} \\
        \hline
        \multirow{4}{=}{\textbf{Fine-tuned}} & agenda-BERT  & \textbf{0.70} & 0.61 & 0.66 & 0.02 & 0.02 & 0.02 \\
                                             \cline{2-8}
                                             & agenda-mBERT & 0.68 & 0.67 & 0.68 & 0.03 & 0.04 & 0.04 \\
                                             \cline{2-8}
                                             & agenda-T5    & 0.38 & 0.38 & 0.39 & 0.06 & 0.05 & 0.06 \\
                                             \cline{2-8}
                                             & agenda-mT5   & 0.33 & 0.34 & 0.34 & 0.03 & 0.04 & 0.04 \\
        \hline
        \multirow{4}{=}{\textbf{Pre-trained +Fine-tuned}} & agenda-rte-en-BERT  & 0.68 & 0.63 & 0.66 & 0.03 & 0.03 & 0.01 \\
                                                          \cline{2-8}
                                                          & agenda-rte-bi-mBERT & 0.66 & 0.68 & 0.67 & 0.03 & 0.04 & 0.02 \\
                                                          \cline{2-8}
                                                          & agenda-rte-en-T5    & 0.68 & 0.64 & 0.66 & 0.02 & 0.03 & 0.02 \\
                                                          \cline{2-8}
                                                          & agenda-rte-bi-mT5   & \textbf{0.70} & \textbf{0.73} & \textbf{0.71} & 0.03 & 0.04 & 0.03 \\                                
        \hline
    \end{tabular}
    \caption{Trained textual entailment evaluation results. We report averaged F1-scores across three runs, and their standard deviation. An "m" in the models name indicates the use of the multi-lingual underlying model, while "rte-en" and "rte-bi" refer to the use of English-only or bi-lingual pre-training using our combined RTE dataset.}
    \label{tab:te_results}
\end{table*}

\subsection{Textual Entailment Agenda Detection}
Our proposed model leverages the power of textual entailment by combining general RTE pre-training and agenda-specific fine-tuning to robustly detect agendas in short-form social media messages. In addition to our main model, we also train and evaluate BERT models and compare against variants that do not include the RTE pre-training.

Fine-tuning our models on the agenda data while following the textual entailment framework resulted in the highest performing models, as can be seen in the detailed results presented in Table \ref{tab:te_results}. In a multilingual setting, the mT5 model, which was pre-trained for textual entailment on our bi-lingual combined RTE dataset (with 30\% of the examples translated into French) and then fine-tuned on the agenda data, outperforms all other baselines, including the conventional multi-label multi-class classification techniques. However, for English-only scenarios, direct fine-tuning of BERT achieves comparable results. 

This highlights that pre-training the model on the combined RTE dataset has a major impact on performance, as it gives the model strong task-specific knowledge, allowing it to tackle the textual entailment problem with ease. By fine-tuning on in-domain data, we observe even better results.

\begin{table*}[hbt!]
    \centering
    \begin{tabular}{p{0.13\linewidth} | p{0.13\linewidth} | p{0.05\linewidth} | p{0.05\linewidth} | p{0.07\linewidth} || p{0.05\linewidth} | p{0.05\linewidth} | p{0.07\linewidth}}
        \hline
        \multicolumn{2}{c |}{\multirow{2}{*}{\textbf{Models}}} & \multicolumn{3}{c ||}{\textbf{AVG}} & \multicolumn{3}{ c }{\textbf{STDEV}} \\ 
        \cline{3-8}
        \multicolumn{2}{c |}{} & \textbf{EN} &\textbf{FR} & \textbf{Overall} & \textbf{EN} & \textbf{FR} & \textbf{Overall} \\
        \hline
        \multirow{6}{=}{\textbf{Fine-tuned}} & tfidf-SVM & 0.45 & 0.47 & 0.46 & 0.14 & 0.13 & 0.13 \\
                                             \cline{2-8}
                                             & mlc-BERT  & 0.48 & 0.36 & 0.43 & 0.04 & 0.05 & 0.03 \\
                                             \cline{2-8}
                                             & mlc-mBERT & \textbf{0.53} & \textbf{0.45} & \textbf{0.50} & 0.03 & 0.04 & 0.03 \\
                                             \cline{2-8}
                                             & mlc-T5    & 0.35 & 0.34 & 0.35 & 0.06 & 0.03 & 0.05 \\
                                             \cline{2-8}
                                             & mlc-mT5   & 0.36 & 0.35 & 0.36 & 0.04 & 0.04 & 0.04 \\
        \hline                          
    \end{tabular}
    \caption{Evaluation results for models trained via the traditional multi-class multi-label classification approach. We report averaged F1-scores across three runs, and their standard deviation. An "m" in the models name indicates the use of the multi-lingual underlying model, and "mlc" stands for Multi-label Classification.}
    \label{tab:mlc_results}
\end{table*}

\subsection{Conventional Text Classification for Agenda Detection}
To detect agendas in social media, we also explore traditional text classification techniques. To this end, we train classifiers for the multi-label, multi-class task using Support Vector Machines (SVM) \citep{boser1992training}. The textual features are extracted through the TF-IDF vectorization yielding 1024 features. Additionally, we deploy BERT and T5 models as baselines, trained for sequence classification. Given T5's text-to-text architecture, we train the model using the tweet message as the source sequence and as the target sequence we use the agenda labels, represented as comma-delimited strings. During testing, T5 generates up to 32 tokens for a given text, which are then parsed and converted into agenda predictions.

The results from this experiment are presented in Table \ref{tab:mlc_results}. Only slightly better than our zero-shot baselines, we see relatively low performance scores. In a surprising turn of events, the Multi-label Classification (MLC) BERT-based models showed a significant decline in performance compared to their Textual Entailment counterparts, agenda-BERT and agenda-mBERT, despite being trained on the same data. In fact, agenda-BERT shows statistically significant improvement over mlc-BERT and mlc-mBERT at significance level $\alpha$ = 0.01, while the same can be observed for agenda-mBERT at at significance level $\alpha$ = 0.05.

Our hypothesis is that by exploiting BERT’s architecture, which is inherently suited for textual entailment, led to agenda-BERT’s superior performance. The limited size of our agenda training data, which has far fewer examples than what is typically required to produce robust models, could also have contributed to the sub-optimal results. 

\subsection{Discussion}
All models trained for agenda under the Textual Entailment framework preform statistically significantly better ($\alpha$ = 0.05) than all the 0-shot pre-trained models (model's name stating with "rte"), including mnli-BART and Semantic Search models. Also, they significantly outperform all of the MLC models at significance level $\alpha$ = 0.05. Restricting the significance level to $\alpha$ = 0.01, we observe that only our best performing model (agenda-rte-bi-mT5) still performs statistically significantly better than all of the 0-shot and MLC models. 

\begin{figure}[hbt!]
    \centering
    \includegraphics[width=0.45\textwidth]{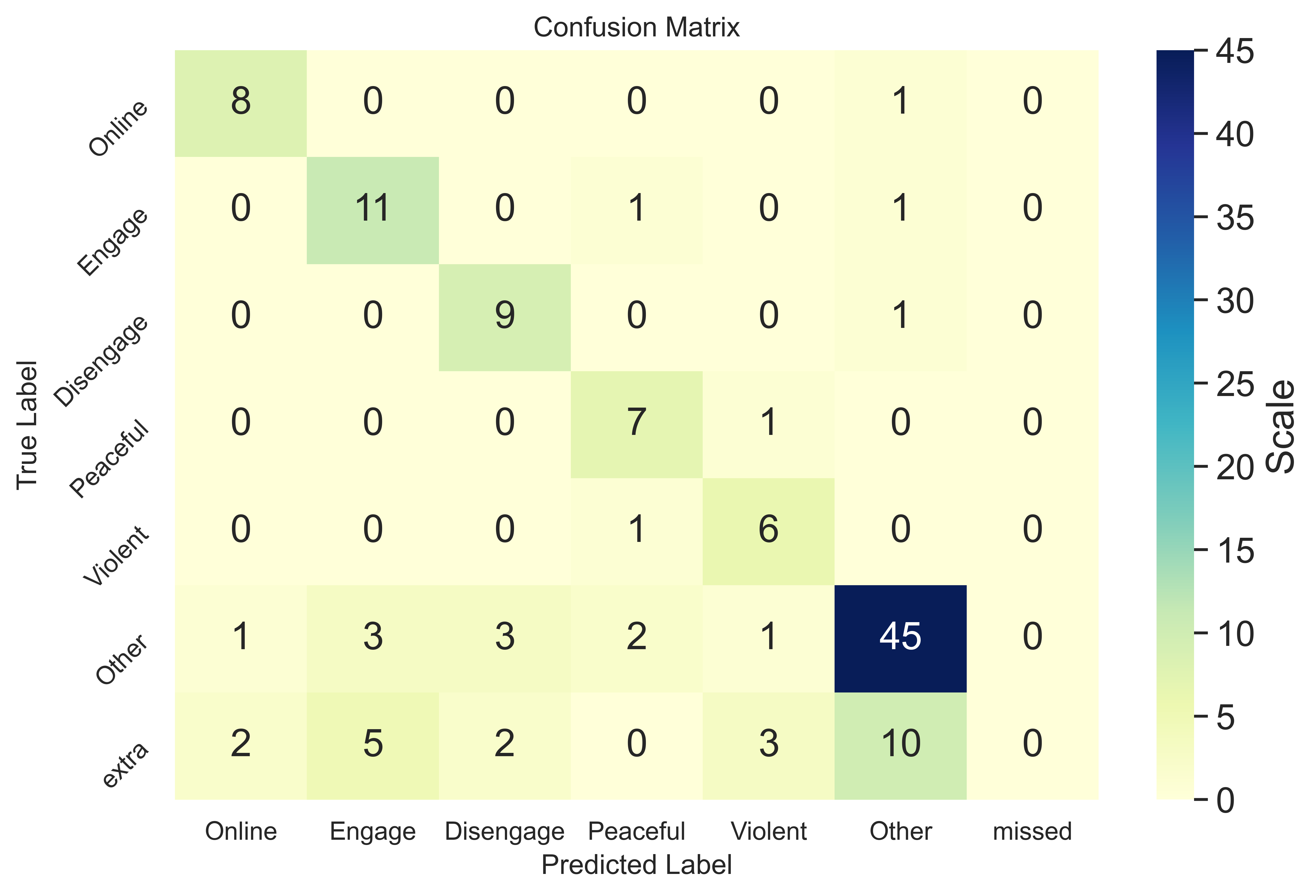}
    \caption{Confusion matrix of agenda-rte-bi-mT5 French results. The extra labels are in the bottom row and the missed labels are in the rightmost column.}
    \label{fig:conf_matrix}
\end{figure}

To gain insight into the limitations of our model and identify areas that may require improvement, we calculate the confusion matrix based on the predictions generated by our best-performing model (agenda-rte-bi-mT5) on the French test set (Figure \ref{fig:conf_matrix}). The multi-class multi-label nature of the task yields false positives, false negatives, introduces the presence of extra labels, where a false positive does not have a corresponding false negative, and missed labels, where a false negative does not have a corresponding false positive.

The agenda-rte-bi-mT5 model tends to over-label, resulting in more predicted labels than actual true labels, with the exception of "Peaceful Protest". Most of the excessive predictions for "Online Solidarity", "Engagement", "Disengagement", and "Violent Action" fall under the "Other" class, while the extra predictions for "Other" are dispersed among all other classes. However, this can be improved by using a better stated "Other" hypothesis or by including of out-of-domain "Other" messages in the training dataset. The example shown in Figure \ref{fig:conf_matrix} does not have any missed labels, however, in the Appendix \ref{sec:appendix} we include cases that do. Missed labels occur when the model lacks confidence in assigning any label to the input message.

Our model incorrectly classified some instances of "Violent Action" as "Peaceful Protest". One such message, translated into English, states \textit{"... to get what we want, peaceful demonstrations are no use! You have to do as in Corsica or as in the suburbs!!!"}. We believe that by integrating external knowledge into the textual entailment process, e.g. knowledge regarding the violent incidents that occurred in Corsica during the French Election of 2022, could lead to further improvement in the model's performance.

\section{Conclusion}
The methodology we have presented for detecting agendas with limited or non-existent labeled examples demonstrates that it is possible to overcome the need for a vast amount of annotated data. This is evident from the superior evaluation results observed by our models. Through an extensive evaluation of various techniques and approaches applied to a small corpus of annotated Twitter messages centered on the 2022 French Presidential Elections, we have shown that treating the task of text classification as a textual entailment problem produces promising results that could not have been achieved through equivalent conventional sequence classification methods.

Our proposed model offers the advantage of not being limited to a set of predefined labels and allows for the testing of an arbitrary number of hypotheses to uncover a multitude of agendas. This versatility makes it an effective tool in detecting new and emerging influence campaigns in social media. However, the spread of agendas is not limited to just Twitter and can also occur through other media such as news articles and blogs, leading us to the next step of studying the applicability of our techniques in longer forms of text and discovering what new insights can be learned.

Our study demonstrated that the textual entailment framework is versatile and can be effectively trained to yield competitive and reasonable results in more than one languages. Our processing pipeline can support a wide range of languages, provided that there are robust pre-trained machine translation models available. Interestingly, our findings suggest that exact translations are not necessary for the models to be able to capture the relationships between premises and hypotheses. This makes our approach flexible and adaptable to a multi-lingual environment. 

\section{Acknowledgements}
This paper is based upon work supported by the Defense Advanced Research Projects Agency (DARPA) under Contract No. HR001121C0186. Any opinions, findings and conclusions or recommendations expressed in this material are those of the authors and do not necessarily reflect the views of DARPA or the U.S. Government.

\section{Ethical Considerations and Limitations}
In our research, we remain mindful of potential biases in the data sources and the semi-automated labeling processes, as well as the impact of our models’ outputs. For instance, our data sources, which include the MNLI and SNLI corpora used to fine-tune our models, may not be representative of the general population or the target domain. The MNLI corpus does not cover all domains and may be skewed towards certain genres. It may also contain annotation artifacts or biases, such as lexical or syntactic cues, that could influence the difficulty of the task for some models. Similarly, the SNLI corpus lacks an annotation manual or guidelines, potentially leading to inconsistent or subjective judgments by the annotators. Furthermore, both corpora may suffer from indeterminacies of event and entity coreference, which could affect the interpretation and labeling of sentence pairs. Additionally, our semi-automated labeling processes may introduce noise or errors.

Additionally, it is essential to acknowledge that our findings are subject to the limitations posed by the size of our agenda training data and the constraints associated with using zero-shot and few-shot learning. Specifically, we note that the size of our agenda training data is relatively small compared to the large-scale datasets used for pre-training language models, which may limit the generalization and robustness of our models. The constraints associated with using zero-shot and few-shot learning are related to the dependency on the quality and relevance of the natural language hypotheses, the difficulty of generating diverse and informative hypotheses, and the challenge of evaluating the reliability of the models.

In the course of our experiments, we recognize the importance of ethical considerations and acknowledge certain limitations inherent to our approach. We have implemented measures to ensure responsible and respectful data collection, annotation, and model development. These measures adhere to the best practices and guidelines for data annotation and quality assurance, including the use of multiple annotators, conflict resolution, and feedback provision.

In our model development and evaluation, we apply principles of fairness and accountability. This involves clearly and explicitly defining the problem and the objectives of the model. We collect and analyze data relevant to the problem and objectives, ensuring it is properly labeled, cleaned, and balanced. The selection and implementation of algorithms and techniques are tailored to suit the problem and objectives. We evaluate and validate the model's performance and behavior using appropriate metrics and methods, as well as comparing the results with baseline models. Finally, we use tools and techniques to explain and interpret the model’s outputs and decisions.

Despite these measures, we remain aware of potential biases in our data sources and semi-automated labeling processes, as well as the impact of our models’ outputs. We acknowledge that our data sources may not fully represent the general population or the target domain, and our semi-automated labeling processes may introduce noise or errors.

Furthermore, we recognize that our findings are subject to limitations due to the size of our agenda training data and the constraints of zero-shot and few-shot learning. Our agenda training data is relatively small compared to the large-scale datasets used for pre-training language models, which may limit our models' generalization and robustness. The constraints of zero-shot and few-shot learning relate to the quality and relevance of the natural language hypotheses, the challenge of generating diverse and informative hypotheses, and the difficulty of evaluating the models' reliability.

\section{Bibliographical References}
\bibliographystyle{lrec-coling2024-natbib}
\bibliography{references}

\section{Appendix}\label{sec:appendix}

\subsection{Appendix A: Alternative NLI Datasets} \label{sec:appendix_a}
In this section of the appendix, we discuss alternative datasets that exist for NLI tasks. The XNLI corpus that has French NLI examples. According to \cite{conneau2018xnli}, the XNLI corpus contains 5000 test and 2500 developent pairs sourced from the MNLI corpus, automatically translated into 14 languages, including French. Therefore, there are a total of 7500 French instances in the XNLI corpus. Since we are using MNLI and automatically translate 30\% of instances into French (approximately 100K instances), we decided not to use XNLI.

The QNLI corpus \cite{wang2018glue} is a collection of 108k sentence pairs that are automatically derived from the Stanford Question Answering Dataset. If we were to use this dataset to pre-train our models, we could experiment by phrasing our hypothesis as questions, rather than statements.

The ANLI \cite{nie-etal-2020-adversarial} has 169k sentence pairs that are adversarially written by humans to challenge NLI models. This could be useful for our model to learn more detailed entailment relationships, further boosting its ability to adapt to our task. We did not consider ANLI, as our combined RTE dataset already exceeded 1M training examples.

\subsection{Appendix B: Agenda Dataset \& Human Annotation Process} \label{sec:appendix_b}
This section of the Appendix outlines the process of creating the agenda dataset, which is based on a collection of tweets pertaining to the 2022 French Presidential Elections. The dataset comprises messages gathered from Twitter between November 12, 2021, and April 30, 2022. Although Twitter supports multilingual content, the collected messages were primarily in French, filtered using keywords that included the names of the candidates and their official Twitter accounts. It's important to note that the original dataset did not include agenda annotations.

\paragraph{Dataset Creation for Agenda Detection in Tweets}
The initial Twitter data consisted of a substantial collection of 17.8 million unique texts distributed across 21 files. An examination of a sample of tweets from the final dataset confirmed French as the dominant language.

To create a manageable and representative subset, we took a 10\% random sample from each file. The pre-processing steps involved the removal of extra white-space and newlines to clean the data. Given that the tweets were collected before 2023, the maximum length of each message was restricted to 280 characters, in line with Twitter's character limit at the time. This comprehensive process ensured the creation of a clean, representative, and manageable dataset for our study.

\paragraph{Bootstrapping and Human-in-the-Loop Annotation}
For this study, we addressed the lack of agenda labels in the raw tweets by employing a bootstrapping technique. This involved the use of a multi-lingual sentence embedding model\footnote{Multi-lingual sentence embedding model: \url{https://huggingface.co/sentence-transformers/paraphrase-multilingual-mpnet-base-v2}} \citep{reimers-2020-multilingual-sentence-bert}, which converts text data into numerical representations that encapsulate semantic similarities. The model was utilized to compute the cosine similarity between the embeddings of the agenda definitions (referenced in Table \ref{tab:definitions} of the main paper) and the embeddings of each tweet.

To accommodate the French language, machine translation was used to generate French versions of the pre-defined agenda definitions via the Google Translation API. 

We implement an automatic labeling procedure for tweets by ranking sampled messages using cosine similarity scores computed between message embeddings and agenda definition embeddings. This process automatically assigns the top 500 highest-ranking messages for each label to their respective agenda classes, while lower-ranked messages are not labeled and are consequently discarded.

Subsequently, two human annotators, who were also authors of this paper, examined the pre-labelled tweets to verify their alignment with the assigned agendas. They reassigned labels when necessary and designated any messages that could not be categorized into any agenda class as "Other".

Upon completing the initial round of annotations, we observed that certain labels had less than 100 messages. To address this imbalance, we proceeded with a follow-up round of bootstrapping and annotation, focusing on the challenging classes of Online Solidarity, Peaceful Protest, and Violent Action.

The annotators brought different linguistic skills to the task. Annotator 1, while not a native speaker of either English or French, was fluent in English and had experience annotating English content in other projects. Annotator 2, a native English speaker with a C1 level of French language proficiency, was able to refer to the original French version of the messages when necessary.

The annotation process, which spanned several weeks, also involved instances of disagreement. These were resolved through discussion until consensus was reached. Messages that remained ambiguous or defied classification were removed from the final dataset, resulting in an Inter-Rater Reliability (IAA) score of 97.5\%, indicating a high level of agreement between the annotators.

\paragraph{Final Dataset Characteristics}
Our final dataset comprises 1012 messages, of which 35 have been assigned multiple labels, indicating that these messages pertain to more than one agenda. The dataset includes both English and French tweets, with a distribution of 10 original English tweets and 1002 original French tweets. To ensure linguistic consistency, all English messages were translated into French and vice versa, using the Google Translate API. Despite the fluency of the annotators potentially mitigating some translation issues, the possibility of error due to machine translation remains. 

To facilitate the training and evaluation of our models, we establish three training, development (dev) and testing sets, such that the dev and test sets are non-overlapping. We create the dev and test sets by taking a 10\% random sample of the total number of messages.

Post-annotation, the number of messages per class varied, ranging from 96 to 120, as detailed in Total column of Table \ref{tab:split}. The "Other" class, serving as a catch-all category for messages that could not be definitively assigned to any of the pre-established agenda labels, contained 506 messages, equivalent to the sum of the other five classes. We acknowledge the limitations of the "Other" category and are committed to refining the dataset in future iterations, which may include exploring alternative definitions for the "Other" category.

\subsection{Appendix C: Detailed Evaluation Results} \label{sec:appendix_c}
Here we present the detailed evaluation results across the three runs of our experiments. In the main section of the paper we only presented the averaged results due to space limitations. As explained in Appendix \ref{sec:appendix_b}, we create three independent train/dev/test splits. We call the collection of models trained and evaluated on a train/dev/test split, a "run". Therefore, our experiments have a total of three "runs", indicated as R1, R2 and R3. 

After training a model on a training set, we perform the evaluation on the corresponding test set by varying the decision threshold for each model. The decision threshold value reflects the model's confidence level, with higher values indicating greater confidence and lower values indicating that predictions with low confidence are accepted, which can result in increased False Positives. We set the minimum possible threshold to 0.3, as we are not interested in trivial scenarios where predictions include all of the available agenda labels. This threshold adjustment is based on the weighted average F1-score. Table \ref{tab:full_results} reports the weighted-average F1-scores for each run, along with the optimal threshold.

\begin{table*}[htb!]
\centering
\resizebox{\textwidth}{!}{%
\begin{tabular}{ccc|cccc|cccc|cccc|}
\cline{4-15}
 &
   &
   &
  \multicolumn{4}{c|}{R1} &
  \multicolumn{4}{c|}{R2} &
  \multicolumn{4}{c|}{R3} \\ \cline{4-15} 
 &
   &
   &
  \multicolumn{1}{c|}{En} &
  \multicolumn{1}{c|}{Fr} &
  \multicolumn{1}{c|}{Overall} &
  Thresh &
  \multicolumn{1}{c|}{En} &
  \multicolumn{1}{c|}{Fr} &
  \multicolumn{1}{c|}{Overall} &
  Thresh &
  \multicolumn{1}{c|}{En} &
  \multicolumn{1}{c|}{Fr} &
  \multicolumn{1}{c|}{Overall} &
  Thresh \\ \hline
\multicolumn{1}{|c|}{\multirow{4}{*}{Semantic Search}} &
  \multicolumn{1}{c|}{\multirow{4}{*}{0-shot}} &
  hypotheses-SBERT &
  \multicolumn{1}{c|}{0.25} &
  \multicolumn{1}{c|}{0.42} &
  \multicolumn{1}{c|}{0.38} &
  0.30 &
  \multicolumn{1}{c|}{0.21} &
  \multicolumn{1}{c|}{0.42} &
  \multicolumn{1}{c|}{0.36} &
  0.31 &
  \multicolumn{1}{c|}{0.18} &
  \multicolumn{1}{c|}{0.45} &
  \multicolumn{1}{c|}{0.38} &
  0.35 \\ \cline{3-15} 
\multicolumn{1}{|c|}{} &
  \multicolumn{1}{c|}{} &
  hypotheses-mSBERT &
  \multicolumn{1}{c|}{0.25} &
  \multicolumn{1}{c|}{0.39} &
  \multicolumn{1}{c|}{0.33} &
  0.30 &
  \multicolumn{1}{c|}{0.21} &
  \multicolumn{1}{c|}{0.38} &
  \multicolumn{1}{c|}{0.31} &
  0.30 &
  \multicolumn{1}{c|}{0.27} &
  \multicolumn{1}{c|}{0.36} &
  \multicolumn{1}{c|}{0.32} &
  0.30 \\ \cline{3-15} 
\multicolumn{1}{|c|}{} &
  \multicolumn{1}{c|}{} &
  labels-SBERT &
  \multicolumn{1}{c|}{0.18} &
  \multicolumn{1}{c|}{0.08} &
  \multicolumn{1}{c|}{0.14} &
  0.30 &
  \multicolumn{1}{c|}{0.16} &
  \multicolumn{1}{c|}{0.14} &
  \multicolumn{1}{c|}{0.16} &
  0.30 &
  \multicolumn{1}{c|}{0.16} &
  \multicolumn{1}{c|}{0.03} &
  \multicolumn{1}{c|}{0.12} &
  0.32 \\ \cline{3-15} 
\multicolumn{1}{|c|}{} &
  \multicolumn{1}{c|}{} &
  labels-mSBERT &
  \multicolumn{1}{c|}{0.20} &
  \multicolumn{1}{c|}{0.21} &
  \multicolumn{1}{c|}{0.21} &
  0.38 &
  \multicolumn{1}{c|}{0.18} &
  \multicolumn{1}{c|}{0.19} &
  \multicolumn{1}{c|}{0.18} &
  0.32 &
  \multicolumn{1}{c|}{0.15} &
  \multicolumn{1}{c|}{0.14} &
  \multicolumn{1}{c|}{0.15} &
  0.38 \\ \hline
\multicolumn{1}{|c|}{\multirow{13}{*}{Textual Entailment}} &
  \multicolumn{1}{c|}{\multirow{5}{*}{0-shot}} &
  rte-en-T5 &
  \multicolumn{1}{c|}{0.48} &
  \multicolumn{1}{c|}{0.46} &
  \multicolumn{1}{c|}{0.48} &
  0.31 &
  \multicolumn{1}{c|}{0.47} &
  \multicolumn{1}{c|}{0.46} &
  \multicolumn{1}{c|}{0.47} &
  0.34 &
  \multicolumn{1}{c|}{0.49} &
  \multicolumn{1}{c|}{0.47} &
  \multicolumn{1}{c|}{0.49} &
  0.30 \\ \cline{3-15} 
\multicolumn{1}{|c|}{} &
  \multicolumn{1}{c|}{} &
  rte-bi-mT5 &
  \multicolumn{1}{c|}{0.39} &
  \multicolumn{1}{c|}{0.42} &
  \multicolumn{1}{c|}{0.41} &
  0.30 &
  \multicolumn{1}{c|}{0.36} &
  \multicolumn{1}{c|}{0.43} &
  \multicolumn{1}{c|}{0.40} &
  0.32 &
  \multicolumn{1}{c|}{0.40} &
  \multicolumn{1}{c|}{0.47} &
  \multicolumn{1}{c|}{0.44} &
  0.33 \\ \cline{3-15} 
\multicolumn{1}{|c|}{} &
  \multicolumn{1}{c|}{} &
  rte-en-BERT &
  \multicolumn{1}{c|}{0.37} &
  \multicolumn{1}{c|}{0.42} &
  \multicolumn{1}{c|}{0.41} &
  0.31 &
  \multicolumn{1}{c|}{0.35} &
  \multicolumn{1}{c|}{0.40} &
  \multicolumn{1}{c|}{0.39} &
  0.47 &
  \multicolumn{1}{c|}{0.41} &
  \multicolumn{1}{c|}{0.46} &
  \multicolumn{1}{c|}{0.45} &
  0.32 \\ \cline{3-15} 
\multicolumn{1}{|c|}{} &
  \multicolumn{1}{c|}{} &
  rte-bi-mBERT &
  \multicolumn{1}{c|}{0.32} &
  \multicolumn{1}{c|}{0.35} &
  \multicolumn{1}{c|}{0.34} &
  0.44 &
  \multicolumn{1}{c|}{0.37} &
  \multicolumn{1}{c|}{0.39} &
  \multicolumn{1}{c|}{0.38} &
  0.39 &
  \multicolumn{1}{c|}{0.41} &
  \multicolumn{1}{c|}{0.42} &
  \multicolumn{1}{c|}{0.42} &
  0.39 \\ \cline{3-15} 
\multicolumn{1}{|c|}{} &
  \multicolumn{1}{c|}{} &
  mnli-BART &
  \multicolumn{1}{c|}{0.33} &
  \multicolumn{1}{c|}{0.38} &
  \multicolumn{1}{c|}{0.38} &
  0.30 &
  \multicolumn{1}{c|}{0.34} &
  \multicolumn{1}{c|}{0.46} &
  \multicolumn{1}{c|}{0.43} &
  0.30 &
  \multicolumn{1}{c|}{0.32} &
  \multicolumn{1}{c|}{0.46} &
  \multicolumn{1}{c|}{0.38} &
  0.30 \\ \cline{2-15} 
\multicolumn{1}{|c|}{} &
  \multicolumn{1}{c|}{\multirow{4}{*}{Fine-Tuned}} &
  agenda-T5 &
  \multicolumn{1}{c|}{0.40} &
  \multicolumn{1}{c|}{0.41} &
  \multicolumn{1}{c|}{0.41} &
  0.30 &
  \multicolumn{1}{c|}{0.32} &
  \multicolumn{1}{c|}{0.32} &
  \multicolumn{1}{c|}{0.32} &
  0.31 &
  \multicolumn{1}{c|}{0.43} &
  \multicolumn{1}{c|}{0.41} &
  \multicolumn{1}{c|}{0.43} &
  0.30 \\ \cline{3-15} 
\multicolumn{1}{|c|}{} &
  \multicolumn{1}{c|}{} &
  agenda-mT5 &
  \multicolumn{1}{c|}{0.32} &
  \multicolumn{1}{c|}{0.32} &
  \multicolumn{1}{c|}{0.32} &
  0.30 &
  \multicolumn{1}{c|}{0.31} &
  \multicolumn{1}{c|}{0.32} &
  \multicolumn{1}{c|}{0.31} &
  0.84 &
  \multicolumn{1}{c|}{0.37} &
  \multicolumn{1}{c|}{0.39} &
  \multicolumn{1}{c|}{0.38} &
  0.78 \\ \cline{3-15} 
\multicolumn{1}{|c|}{} &
  \multicolumn{1}{c|}{} &
  agenda-BERT &
  \multicolumn{1}{c|}{\textbf{0.72}} &
  \multicolumn{1}{c|}{0.60} &
  \multicolumn{1}{c|}{0.66} &
  0.48 &
  \multicolumn{1}{c|}{0.69} &
  \multicolumn{1}{c|}{0.59} &
  \multicolumn{1}{c|}{0.64} &
  0.37 &
  \multicolumn{1}{c|}{0.70} &
  \multicolumn{1}{c|}{0.63} &
  \multicolumn{1}{c|}{0.67} &
  0.53 \\ \cline{3-15} 
\multicolumn{1}{|c|}{} &
  \multicolumn{1}{c|}{} &
  agenda-mBERT &
  \multicolumn{1}{c|}{0.71} &
  \multicolumn{1}{c|}{0.69} &
  \multicolumn{1}{c|}{0.70} &
  0.80 &
  \multicolumn{1}{c|}{0.68} &
  \multicolumn{1}{c|}{0.69} &
  \multicolumn{1}{c|}{0.69} &
  0.49 &
  \multicolumn{1}{c|}{0.65} &
  \multicolumn{1}{c|}{0.64} &
  \multicolumn{1}{c|}{0.64} &
  0.85 \\ \cline{2-15} 
\multicolumn{1}{|c|}{} &
  \multicolumn{1}{c|}{\multirow{4}{*}{Pre-trained \& Fine-Tuned}} &
  agenda-rte-en-T5 &
  \multicolumn{1}{c|}{0.67} &
  \multicolumn{1}{c|}{0.67} &
  \multicolumn{1}{c|}{0.67} &
  0.66 &
  \multicolumn{1}{c|}{0.68} &
  \multicolumn{1}{c|}{0.61} &
  \multicolumn{1}{c|}{0.64} &
  0.78 &
  \multicolumn{1}{c|}{0.70} &
  \multicolumn{1}{c|}{0.63} &
  \multicolumn{1}{c|}{0.66} &
  0.62 \\ \cline{3-15} 
\multicolumn{1}{|c|}{} &
  \multicolumn{1}{c|}{} &
  agenda-rte-bi-mT5 &
  \multicolumn{1}{c|}{0.67} &
  \multicolumn{1}{c|}{\textbf{0.70}} &
  \multicolumn{1}{c|}{0.68} &
  0.41 &
  \multicolumn{1}{c|}{\textbf{0.72}} &
  \multicolumn{1}{c|}{\textbf{0.72}} &
  \multicolumn{1}{c|}{\textbf{0.72}} &
  0.48 &
  \multicolumn{1}{c|}{\textbf{0.71}} &
  \multicolumn{1}{c|}{\textbf{0.77}} &
  \multicolumn{1}{c|}{\textbf{0.74}} &
  0.49 \\ \cline{3-15} 
\multicolumn{1}{|c|}{} &
  \multicolumn{1}{c|}{} &
  agenda-rte-en-BERT &
  \multicolumn{1}{c|}{\textbf{0.72}} &
  \multicolumn{1}{c|}{0.60} &
  \multicolumn{1}{c|}{0.67} &
  0.63 &
  \multicolumn{1}{c|}{0.67} &
  \multicolumn{1}{c|}{0.66} &
  \multicolumn{1}{c|}{0.66} &
  0.47 &
  \multicolumn{1}{c|}{0.66} &
  \multicolumn{1}{c|}{0.63} &
  \multicolumn{1}{c|}{0.65} &
  0.89 \\ \cline{3-15} 
\multicolumn{1}{|c|}{} &
  \multicolumn{1}{c|}{} &
  agenda-rte-bi-mBERT &
  \multicolumn{1}{c|}{0.69} &
  \multicolumn{1}{c|}{0.68} &
  \multicolumn{1}{c|}{\textbf{0.69}} &
  0.79 &
  \multicolumn{1}{c|}{0.64} &
  \multicolumn{1}{c|}{\textbf{0.72}} &
  \multicolumn{1}{c|}{0.68} &
  0.96 &
  \multicolumn{1}{c|}{0.66} &
  \multicolumn{1}{c|}{0.64} &
  \multicolumn{1}{c|}{0.65} &
  0.82 \\ \hline
\multicolumn{1}{|c|}{\multirow{5}{*}{Multi-label   Classification}} &
  \multicolumn{1}{c|}{\multirow{5}{*}{Fine-Tuned}} &
  tfidf-SVM &
  \multicolumn{1}{c|}{0.61} &
  \multicolumn{1}{c|}{0.61} &
  \multicolumn{1}{c|}{0.61} &
  0.31 &
  \multicolumn{1}{c|}{0.38} &
  \multicolumn{1}{c|}{0.38} &
  \multicolumn{1}{c|}{0.38} &
  0.30 &
  \multicolumn{1}{c|}{0.36} &
  \multicolumn{1}{c|}{0.41} &
  \multicolumn{1}{c|}{0.39} &
  0.31 \\ \cline{3-15} 
\multicolumn{1}{|c|}{} &
  \multicolumn{1}{c|}{} &
  mlc-T5 &
  \multicolumn{1}{c|}{0.31} &
  \multicolumn{1}{c|}{0.32} &
  \multicolumn{1}{c|}{0.32} &
  0.30 &
  \multicolumn{1}{c|}{0.32} &
  \multicolumn{1}{c|}{0.32} &
  \multicolumn{1}{c|}{0.32} &
  0.30 &
  \multicolumn{1}{c|}{0.42} &
  \multicolumn{1}{c|}{0.38} &
  \multicolumn{1}{c|}{0.40} &
  0.30 \\ \cline{3-15} 
\multicolumn{1}{|c|}{} &
  \multicolumn{1}{c|}{} &
  mlc-mT5 &
  \multicolumn{1}{c|}{0.34} &
  \multicolumn{1}{c|}{0.32} &
  \multicolumn{1}{c|}{0.33} &
  0.30 &
  \multicolumn{1}{c|}{0.33} &
  \multicolumn{1}{c|}{0.35} &
  \multicolumn{1}{c|}{0.34} &
  0.30 &
  \multicolumn{1}{c|}{0.40} &
  \multicolumn{1}{c|}{0.39} &
  \multicolumn{1}{c|}{0.40} &
  0.30 \\ \cline{3-15} 
\multicolumn{1}{|c|}{} &
  \multicolumn{1}{c|}{} &
  mlc-BERT &
  \multicolumn{1}{c|}{0.51} &
  \multicolumn{1}{c|}{0.32} &
  \multicolumn{1}{c|}{0.43} &
  0.30 &
  \multicolumn{1}{c|}{0.43} &
  \multicolumn{1}{c|}{0.36} &
  \multicolumn{1}{c|}{0.40} &
  0.30 &
  \multicolumn{1}{c|}{0.49} &
  \multicolumn{1}{c|}{0.41} &
  \multicolumn{1}{c|}{0.46} &
  0.30 \\ \cline{3-15} 
\multicolumn{1}{|c|}{} &
  \multicolumn{1}{c|}{} &
  mlc-mBERT &
  \multicolumn{1}{c|}{0.51} &
  \multicolumn{1}{c|}{0.42} &
  \multicolumn{1}{c|}{0.47} &
  0.30 &
  \multicolumn{1}{c|}{0.53} &
  \multicolumn{1}{c|}{0.45} &
  \multicolumn{1}{c|}{0.50} &
  0.30 &
  \multicolumn{1}{c|}{0.56} &
  \multicolumn{1}{c|}{0.49} &
  \multicolumn{1}{c|}{0.53} &
  0.30 \\ \hline
\end{tabular}%
}
\caption{Evaluation results for all models across three runs. We report weighted-average F1-scores across three runs. An "m" in the models name indicates the use of the multi-lingual underlying model, while "rte-en" and "rte-bi" refer to the use of English-only or bi-lingual pre-training using our combined RTE dataset. The "mlc" prefix stands for Multi-label Classification.}
\label{tab:full_results}
\end{table*}

\subsection{Appendix D: Additional Confusion Matrix Results} \label{sec:appendix_d}
\begin{figure}[hbt!]
    \centering
    \includegraphics[width=0.45\textwidth]{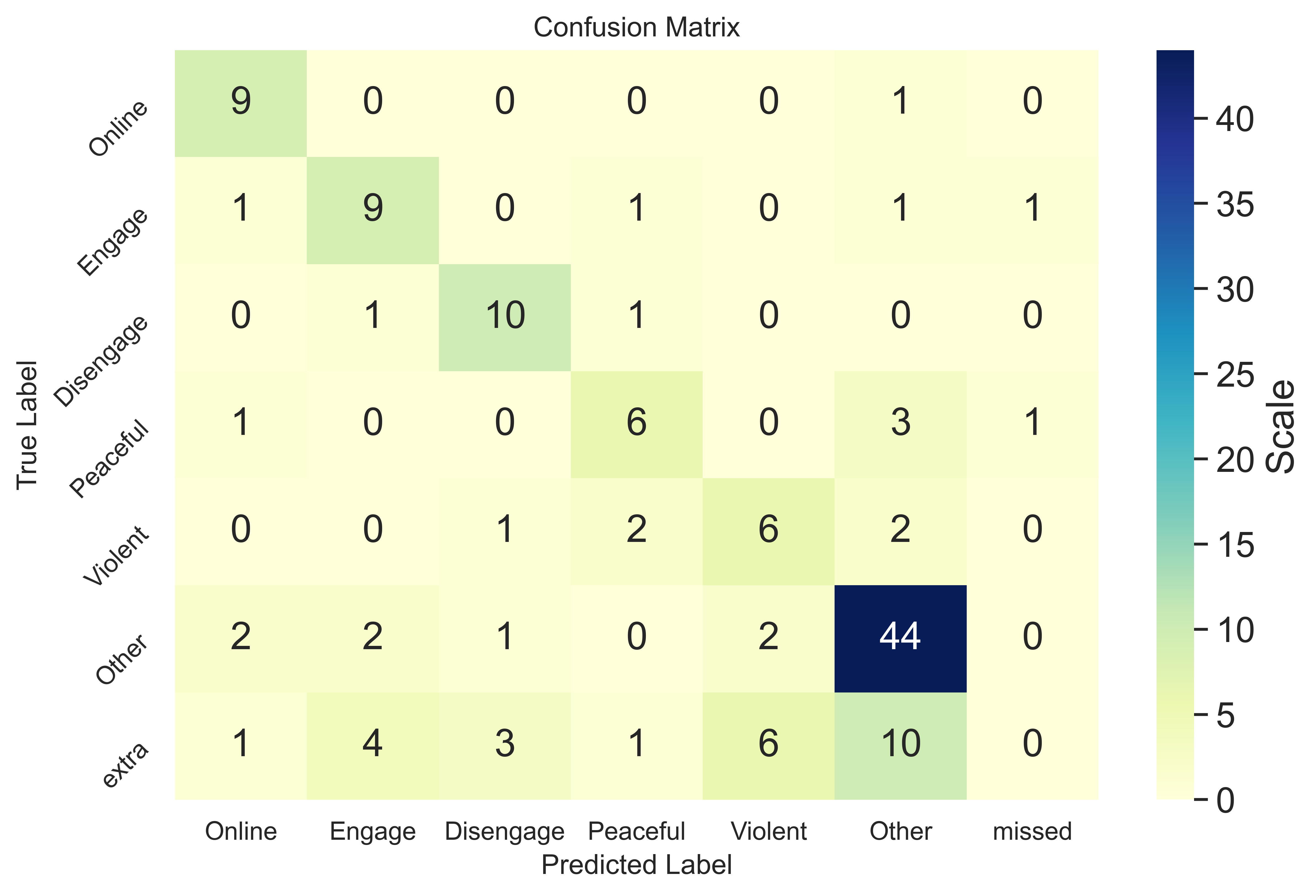}
    \caption{Confusion matrix for agenda-rte-bi-mT5 French results on Run 1. The extra labels are in the bottom row and the missed labels are in the rightmost column.}
    \label{fig:conf_matrix_2}
\end{figure}

In this section of the appendix, we present additional confusion matrix results. The case shown in \ref{fig:conf_matrix_2} contains missed labels, that can occur when the model lacks confidence in assigning any label to the input message.

\end{document}